\documentclass[lettersize,journal]{IEEEtran}
\usepackage{amsmath,amsfonts}
\usepackage{algorithmic}
\usepackage{algorithm}
\usepackage{array}
\usepackage[caption=false,font=normalsize,labelfont=sf,textfont=sf]{subfig}
\usepackage{textcomp}
\usepackage{stfloats}
\usepackage{url}
\usepackage{verbatim}
\usepackage{graphicx}
\usepackage{booktabs}
\usepackage{multirow}
\usepackage{multicol}
\usepackage{color}
\usepackage{hyperref}
\hypersetup{
    pagebackref=true,
	colorlinks=true,
	linkcolor=red,
	filecolor=blue,      
	urlcolor=red,
	citecolor=green,
}
\usepackage{cite}
\begin{document}

\title{FiLo++: Zero-/Few-Shot Anomaly Detection by Fused Fine-Grained Descriptions and Deformable Localization}

\author{Zhaopeng Gu, Bingke Zhu, Guibo Zhu, Yingying Chen, Ming Tang,  \IEEEmembership{Member,~IEEE}, Jinqiao Wang,\\ \IEEEmembership{Member,~IEEE}

\thanks{Zhaopeng Gu and Ming Tang are with the Foundation Model Research Center, Institute of Automation, Chinese Academy of Sciences, Beijing 100190, China, and also with the School of Artifcial Intelligence, University of Chinese Academy of Sciences, Beijing 100049, China (e-mail: guzhaopeng2023@ia.ac.cn; tangm@nlpr.ia.ac.cn).}
\thanks{Bingke Zhu and Yingying Chen are at the Foundation Model Research Center, Institute of Automation, Chinese Academy of Sciences, Beijing 100190, China (e-mail: bingke.zhu@nlpr.ia.ac.cn; yingying.chen@nlpr.ia.ac.cn).}
\thanks{Guibo Zhu is with the Foundation Model Research Center, Institute of Automation, Chinese Academy of Sciences, Beijing 100190, China, also with the School of Artifcial Intelligence, University of Chinese Academy of Sciences, Beijing 100049, China, and also with the Shanghai Artificial Intelligence Laboratory, Shanghai 200232, China (e-mail: gbzhu@nlpr.ia.ac.cn).}
\thanks{Jinqiao Wang is with the Foundation Model Research Center, Institute of Automation, Chinese Academy of Sciences, Beijing 100190, China, also with the School of Artifcial Intelligence, University of Chinese Academy of Sciences, Beijing 100049, China, also with the Wuhan AI Research, Wuhan 430073, China, also with the Peng Cheng Laboratory, Shenzhen 518066, China, and also with the Guangdong Provincial Key Laboratory of Intellectual Property \& Big Data, Guangdong Polytechnic Normal University, Guangzhou 510665, China (e-mail: jqwang@nlpr.ia.ac.cn).}
}

\markboth{Journal of \LaTeX\ Class Files, VOL. XX, NO. XX, XXXX}%
{Gu \MakeLowercase{\textit{et al.}}: FiLo++: Zero-/Few-Shot Anomaly Detection by Fused Fine-Grained Descriptions and Deformable Localization}


\maketitle
\begin{abstract}
Anomaly detection methods typically require extensive normal samples from the target class for training, limiting their applicability in scenarios that require rapid adaptation, such as cold start. Zero-shot and few-shot anomaly detection do not require labeled samples from the target class in advance, making them a promising research direction. Existing zero-shot and few-shot approaches often leverage powerful multimodal models to detect and localize anomalies by comparing image-text similarity. However, their handcrafted generic descriptions fail to capture the diverse range of anomalies that may emerge in different objects, and simple patch-level image-text matching often struggles to localize anomalous regions of varying shapes and sizes. To address these issues, this paper proposes the FiLo++ method, which consists of two key components. The first component, \textbf{Fus}ed Fine-Grained \textbf{Des}criptions~(FusDes), utilizes large language models to generate anomaly descriptions for each object category, combines both fixed and learnable prompt templates and applies a runtime prompt filtering method, producing more accurate and task-specific textual descriptions. The second component, \textbf{Def}ormable \textbf{Loc}alization~(DefLoc), integrates the vision foundation model Grounding DINO with position-enhanced text descriptions and a Multi-scale Deformable Cross-modal Interaction~(MDCI) module, enabling accurate localization of anomalies with various shapes and sizes. In addition, we design a position-enhanced patch matching approach to improve few-shot anomaly detection performance. Experiments on multiple datasets demonstrate that FiLo++ achieves significant performance improvements compared with existing methods. Code will be available at \href{https://github.com/CASIA-IVA-Lab/FiLo}{https://github.com/CASIA-IVA-Lab/FiLo}.
\end{abstract}


\section{Introduction}
\IEEEPARstart{A}{nomaly} detection is a highly practical task that finds wide application across diverse fields, including detecting product defects in industrial manufacturing~\cite{zhu2024adformer,jiang2024fabgpt,zhu2024pixel, liu2024unistad, yao2023learning}, identifying lesions in medical contexts~\cite{bao2024bmad, huang2024adapting}, and monitoring abnormal behaviors of vehicles and pedestrians in transportation~\cite{chen2025unveiling, liang2024text, zhong2022bidirectional,fang2025amita}. Traditional anomaly detection methods~\cite{roth2022towards, defard2021padim, you2022unified} typically regard the problem as one-class classification, where the model is trained on a large number of normal samples and subsequently attempts to detect out-of-distribution anomalies. Although these methods perform well, they lose effectiveness in scenarios where large-scale normal data collection is challenging (\textit{e.g.}, cold-start settings). Consequently, zero-shot and few-shot anomaly detection methods~\cite{jeong2023winclip,huang2022registration,gu2024filo}, which do not require prior data from the target category, have gained considerable attention. In these methods, only a small number of normal samples are optionally provided as references during testing, and the methods can detect object categories that have never been encountered during training.

\begin{figure}[!t]
\centering
\includegraphics[width=\linewidth]{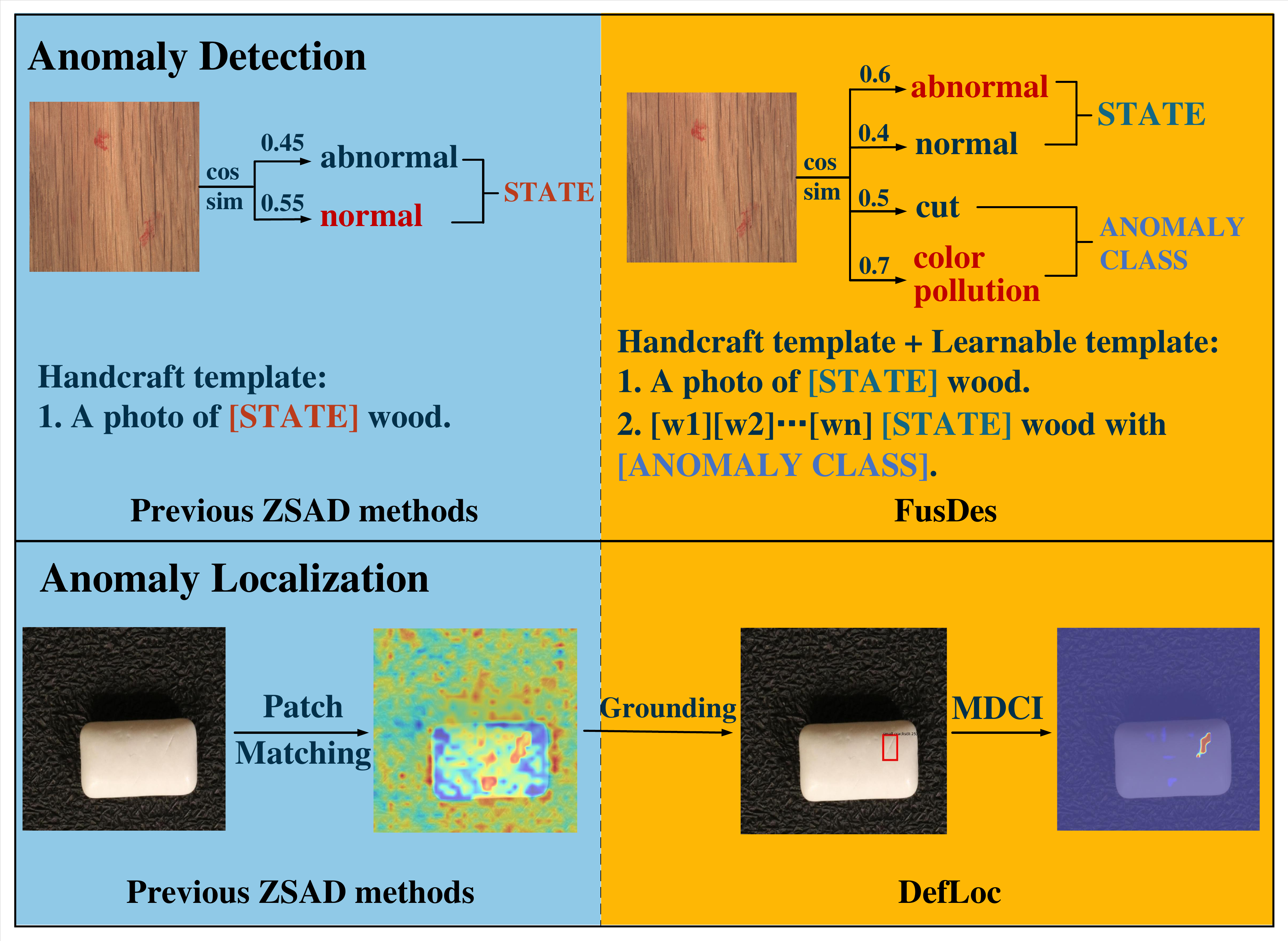}
\caption{Comparison of anomaly detection and localization between FiLo++ and previous ZSAD methods. Previous ZSAD methods utilize generic anomaly descriptions, which may lead to errors. Our FusDes enhances detection accuracy by fine-grained anomaly descriptions, learnable templates, and runtime prompt filtering. For localization, existing ZSAD methods typically compare image patches directly with text features, resulting in false positives in background regions. Our DefLoc method effectively eliminates background areas and improves localization accuracy by employing Grounding DINO, position-enhanced text descriptions, and the MDCI module.}
\label{fig_1}
\end{figure}

Existing zero-shot and few-shot anomaly detection approaches~\cite{jeong2023winclip,huang2022registration,gu2024filo} primarily build on multimodal pre-trained models such as CLIP~\cite{radford2021learning}. Trained on extremely large-scale image-text pair datasets, multimodal models exhibit remarkable zero-shot performance in image classification~\cite{zhou2022learning}, semantic segmentation~\cite{shang2022cross, li2024bidirectional}, and object detection~\cite{liu2025grounding}. Anomaly detection methods based on multimodal pretrained models~\cite{jeong2023winclip,gu2024filo} often rely on handcrafted text prompts conveying ``normal” and ``abnormal” semantics, then determine whether each image patch is anomalous by comparing its feature similarity to text embeddings. Such techniques offer a solution for both Zero-Shot Anomaly Detection~(ZSAD) and Few-Shot Anomaly Detection~(FSAD), yet they face limitations in two main aspects. First, regarding detection, manually crafted general descriptions lack flexibility and fail to capture the diverse types of anomalies across different object categories. Second, for localization, naively matching patch features with text prompts struggles to detect anomalous regions that span multiple patches or have varying shapes.

In response to these challenges, this paper proposes a method called FiLo++, which offers effective solutions for anomaly detection and localization. For detection, we design a \textbf{Fus}ed Fine-Grained \textbf{Des}criptions~(FusDes) module to improve anomaly detection. First, this module leverages the extensive cross-domain knowledge of Large Language Models~(LLMs) to generate specific types of anomalies potentially appearing in the test sample, replacing the generic, manually crafted prompts with content more tailored to each test sample. Second, existing research shows that the template of the text prompt in multimodal models significantly affects performance. Prior methods usually adopt the prompt template from CLIP~\cite{radford2021learning} that is originally designed for ImageNet classification~\cite{deng2009imagenet}, which may not fully suit anomaly detection tasks. To this end, we introduce new prompt templates specialized for anomaly detection and combine fixed, human-designed templates with learnable, adaptive templates to produce text prompts better suited to anomaly detection. Finally, we propose a runtime prompt filtering strategy that boosts the distinguishability between normal and abnormal text features, yielding the final fused fine-grained description. Compared to the ``normal” vs. ``abnormal” general descriptions used by existing methods, the FusDes module, which combines LLM priors, fixed and learnable prompt templates and runtime filtering, substantially enhances anomaly detection capabilities.

For localization, we design a \textbf{Def}ormable \textbf{Loc}alization~(DefLoc) module to overcome difficulties in localizing anomalous regions of various sizes and shapes across multiple patches. This module proceeds in three main steps. First, DefLoc utilizes detailed anomaly information generated by the LLMs and employs the vision foundation model Grounding DINO~\cite{liu2025grounding} for initial anomaly localization. Although Grounding DINO alone shows poor performance in anomaly detection, it effectively filters out backgrounds and homogeneous areas irrelevant to anomalies, thus helping subsequent localization steps. Second, DefLoc integrates the positional information from Grounding DINO’s initial localization results into the text descriptions, making them more accurate. Third, DefLoc applies a Multi-scale Deformable Cross-modal Interaction~(MDCI) module, which aggregates image patch features using deformable convolutions~\cite{zhu2019deformable} at multiple scales and thereby strengthens detection of anomalous regions with different shapes and sizes.

Furthermore, we design a position-enhanced patch matching approach to support few-shot anomaly detection. By leveraging the preliminary localization results from the DefLoc module, we constrain the scope of patch matching to improve detection and localization performance in FSAD scenarios.

FiLo++ is an extension of our work FiLo~\cite{gu2024filo}, which is published in ACM MM 2024. Compared to FiLo, FiLo++ makes three primary improvements: 1) FiLo++ replaces the FG-Des module with the FusDes module that merges fixed prompt templates, learnable templates, and runtime prompt filtering to further enhance the alignment between text and image features.; 2) FiLo++ employs multi-scale deformable convolutions in place of standard convolutions in the DefLoc module to better detect anomalies of different shapes and sizes; 3) FiLo++ adds a FSAD branch that utilizes DefLoc’s initial localization results to refine patch matching, expanding the applicability of the FiLo++. Beyond making structural modifications, we further conduct experiments under various settings and carry out more extensive ablation studies to provide a comprehensive evaluation of our method.

We perform extensive experiments on various datasets such as MVTec-AD~\cite{bergmann2019mvtec} and VisA~\cite{zou2022spot}, and results show that FiLo++ achieves significant improvements in ZSAD and FSAD. For instance, in the zero-shot scenario on VisA, FiLo++ achieves an image-level AUC of 84.5\% and a pixel-level AUC of 96.2\%.

Our contributions can be summarized as follows:
\begin{itemize}
    \item We propose a FusDes approach that leverages cross-domain knowledge in LLMs to generate detailed anomaly descriptions. By combining fixed text prompt templates with learnable templates and applying a runtime prompt filtering method, we produce text features more suitable for anomaly detection, thereby enhancing both accuracy and interpretability.
    
    \item Additionally, we introduce a DefLoc module that integrates preliminary anomaly localization from the Grounding DINO vision foundation model, position-enhanced text descriptions, and a Multi-scale Deformable Cross-modal Interaction~(MDCI) module to more accurately localize anomalies of different sizes and shapes. Furthermore, by incorporating a few-shot anomaly detection branch, FiLo++ can perform both ZSAD and FSAD, improving its flexibility and generalization.

    \item Extensive experiments on multiple datasets show that FiLo++ significantly outperforms baseline methods. Experimental results proves effective for both zero-shot and few-shot anomaly detection and localization, achieving state-of-the-art performance.

\end{itemize}

\section{Related Work}

\subsection{Anomaly Detection}

Anomaly detection traditionally relies on one-class classification requiring extensive normal training samples. To address data scarcity, ZSAD and FSAD have emerged, largely driven by the success of the multimodal pretrained CLIP~\cite{radford2021learning}. Early explorations such as CLIP-AD~\cite{liznerski2022exposing} and WinCLIP~\cite{jeong2023winclip} utilized handcrafted text prompts (e.g., ``normal'' vs. ``abnormal'') and sliding windows for detection. Subsequent improvements like APRIL-GAN~\cite{chen2023april} and AnomalyCLIP~\cite{zhou2023anomalyclip} introduced linear adapters or learnable vectors to refine feature alignment. More recently, advanced frameworks have been proposed to enhance multimodal interaction. For instance, FocusCLIP~\cite{zhao2024focusclip} introduces a bidirectional self-knowledge distillation framework to focus on anomalous regions by mining visual-text discrepancies. PLOVAD~\cite{xu2025plovad} explores prompt tuning specifically for open-vocabulary video anomaly detection, utilizing a temporal module to capture motion irregularities. Beyond defect detection, Surveillance VALU~\cite{yuan2024surveillance} extends multimodal understanding to event analysis in surveillance videos. While these methods demonstrate the power of Vision-Language Models, they often rely on global discrepancies~\cite{zhao2024focusclip} or focus on high-level temporal events~\cite{xu2025plovad,yuan2024surveillance}. In contrast, FiLo++ targets static industrial and medical anomalies by employing LLMs to generate fine-grained descriptions and resolving cross-semantic ambiguity via runtime filtering, offering superior interpretability and precision for micro-level defects.

In the realm of few-shot learning, patch-level feature comparison methods like PatchCore~\cite{roth2022towards} and RegAD~\cite{huang2022registration} are prevalent. Significant progress has also been made in this setting recently. For example, NIGSF~\cite{xing2023normal} utilizes saliency augmentation to guide segmentation, while Reverse Distillation~\cite{wang2025anomaly} employs latent anomaly suppression to prevent the reconstruction of anomalous patterns. Although these reconstruction and segmentation-based methods achieve high performance, they necessitate a training phase. FiLo++ distinguishes itself by targeting the more challenging zero-shot scenarios or extremely low-data regimes (1-4 shots). By leveraging the preliminary localization from our DefLoc module to constrain patch matching, FiLo++ achieves competitive performance without the need for extensive training data, complementing the one-class classifiers.

Regarding anomaly localization, accurate segmentation of defects with varying shapes and sizes remains a challenge. Methods like SAA~\cite{cao2023segment} utilize Grounding DINO~\cite{liu2025grounding} for segmentation, but direct application often leads to false positives due to domain shift. Furthermore, while multi-scale convolutions~\cite{szegedy2015going,ding2021repvgg} effectively extract features at different scales, fixed-shaped kernels lack flexibility for irregular defects. Deformable Convolution Networks~\cite{zhu2019deformable} address this by adapting the grid structure. In FiLo++, we integrate Grounding DINO as an initial filter and propose a Multi-scale Deformable Cross-modal Interaction (MDCI) module. This combination allows our model to robustly localize anomalies of various morphologies while effectively suppressing background noise.

\subsection{Multi-Scale Convolution}
Multi-scale convolution combines convolution kernels of different receptive field sizes, enabling effective feature extraction for objects of various sizes in an image. This approach has demonstrated outstanding performance in numerous vision tasks and has become highly popular in computer vision research. InceptionNet~\cite{szegedy2015going} is a pioneering example that uses 1$\times$1, 3$\times$3, and 5$\times$5 kernels in parallel within the same layer and then concatenates the results along the channel dimension. RepVGG~\cite{ding2021repvgg} decomposes larger convolution kernels into multiple 3$\times$3 kernels, substantially reducing model parameters and improving inference speed. MixConv~\cite{tan2019mixconv} applies different kernel sizes to different channels within the same convolution operation, balancing multi-scale benefits and computational efficiency. Nevertheless, these methods all rely on fixed-shaped kernels, which lack flexibility. Deformable Convolution Network~\cite{zhu2019deformable} introduces deformable convolution operations that greatly enhance feature extraction for objects with irregular shapes, which is highly beneficial for anomaly detection. In FiLo++, we incorporate both multi-scale and deformable convolutions, fully leveraging multimodal image-text features to design the MDCI module, which accurately localizes anomalies of various sizes and shapes.

\begin{figure*}[t]
  \centering
  \includegraphics[width=\textwidth]{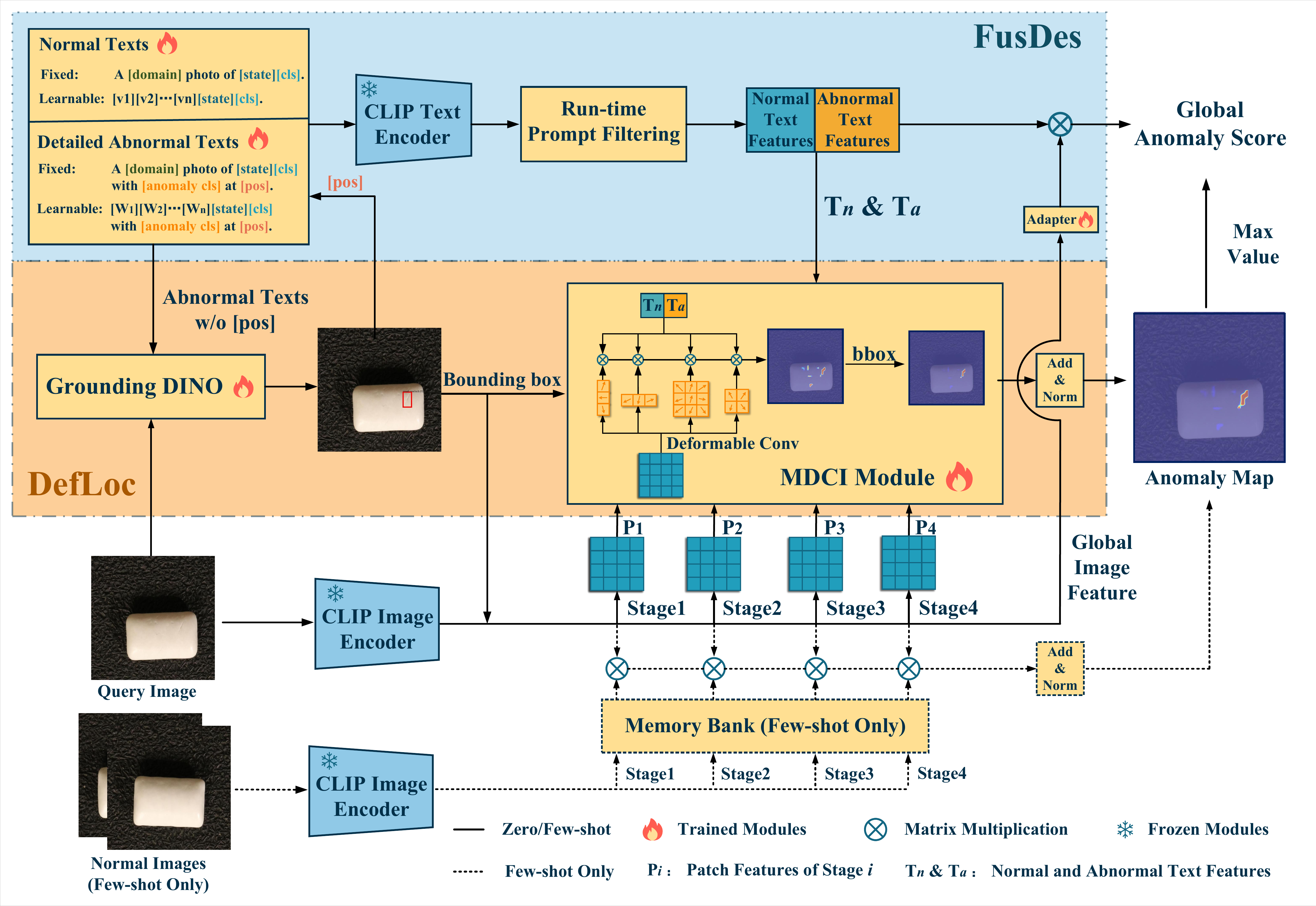} 
  \caption{Overall architecture of FiLo++. Given an input image, an LLM generates fine-grained anomaly types. The normal and detailed anomaly texts are processed by Grounding DINO to obtain bounding boxes, then combined with fixed and learnable templates and encoded by the CLIP Text Encoder with runtime prompt filtering to produce \( T_n \) and \( T_a \). The image’s intermediate patch features interact with the text features through the MDCI module to create the vision-language anomaly map. A few-shot anomaly map is generated using the memory bank of few-shot normal samples. Finally, global image features are compared with the fused text features to obtain the global anomaly score.}
  \label{fig:arch}
\end{figure*}

\section{Method}
\subsection{Overview}

This paper proposes a novel zero-shot and few-shot anomaly detection method, FiLo++, which enhances the performance of anomaly detection and localization through two modules: FusDes and DefLoc. Specifically, for anomaly detection, we design the Fused Fine-Grained Description module (FusDes, Section~\ref{sec:FusDes}), which leverages detailed anomaly descriptions provided by large language models. FusDes combines fixed templates with learnable text vectors and implements a runtime prompt filtering strategy to obtain text features that more accurately match anomaly detection images. The FusDes module not only determines whether an image contains anomalies but also identifies the specific types of anomalies within the image, significantly improving the interpretability of the method. For anomaly localization, we develop the Deformable Localization module (DefLoc, Section~\ref{sec:DefLoc}), which accurately locates anomalies of varying sizes and dimensions through initial localization using Grounding DINO, position-enhanced text descriptions, and a Multi-scale Deformable Cross-modal Interaction module.

\subsection{FusDes}
\label{sec:FusDes}

Numerous existing methods demonstrate that the quality of text prompts significantly affects the performance of anomaly detection based on image-text pre-trained models when inferring new categories. Therefore, we first investigate prompt engineering to generate more precise and efficient text prompts to enhance detection accuracy. The FusDes module consists of three main components: 1) generation of fine-grained anomaly descriptions based on LLMs; 2) combination of fixed templates with learnable text vectors; and 3) runtime prompt filtering. These three components are detailed below.

\subsubsection{Fine-Grained Anomaly Descriptions}

Initial CLIP-based anomaly detection methods use terms like ``abnormal" to represent anomaly semantics, which fail to capture the diversity of anomaly types. WinCLIP expands anomaly description texts by including terms such as ``damaged," ``flaw," and ``defect." However, these descriptions remain broad and cannot accurately describe different anomaly types on various objects. We require more specific and precise anomaly descriptions to match the rich variety of anomalies. Large language models (LLMs), such as GPT-4, trained on vast image and text datasets, possess extensive ``world knowledge" across various domains. We utilize the powerful knowledge of LLMs to generate detailed anomaly types for each test sample, resulting in more accurate fine-grained descriptions than generic terms like ``damaged" or ``abnormal." These detailed descriptions better match the test images, improving detection precision and enabling the determination of specific anomaly contents in the images based on the similarity between text descriptions and image content, thereby enhancing the method's interpretability.

\subsubsection{Combination of Fixed Templates and Learnable Text Vectors}

After methods like WinCLIP achieve excellent performance on multiple anomaly detection datasets, subsequent approaches typically adopt the same text templates used by WinCLIP to construct text prompts. However, the text template \textit{A photo of [class].} used in WinCLIP is primarily derived from templates employed by CLIP for image classification tasks on the ImageNet dataset, which focus on indicating the foreground object category rather than whether the object contains anomalous parts. Therefore, we modify this template by incorporating the fine-grained descriptions generated by LLMs, changing it to \textit{A [domain] photo of [state] [cls] (with [anomaly cls] at [pos])}. Here, \textit{[domain]} represents the object's domain, \textit{[state]} indicates normal or anomalous status, and \textit{[cls]} denotes the object category name. For anomaly descriptions, the template includes \textit{[anomaly cls]} for detailed anomaly content and \textit{[pos]} for anomaly location, categorized into nine positions: top-left, left, bottom-left, top, center, bottom, top-right, right, and bottom-right.

The modified template significantly enhances both performance and interpretability compared to the original. However, manually crafted templates cannot achieve the optimal solution for anomaly detection tasks. Consequently, we introduce learnable adaptive text templates trained with relevant anomaly detection data. These templates adaptively learn text prompts that better distinguish between normal and anomalous samples based on the image's normal and anomalous content. The adaptive normal and anomalous text templates are defined as follows:
\begin{equation*}
\begin{aligned}
T_n =\ &[V_1][V_2] \dots [V_n][state][cls] \\
T_a =\ &[W_1][W_2] \dots [W_n][state][cls]\ \\
    \ &with\ [anomaly\ cls]\ at\ [pos] \\
\end{aligned}
\end{equation*}
where $[V_i]$ and $[W_i]$ are learnable text vectors, and $T_n$ and $T_a$ represent the normal and anomalous text templates.

By inserting the fine-grained anomaly descriptions generated by LLMs into the \textit{[anomaly cls]} field of the adaptive text templates, we obtain complete text prompts. These fine-grained anomaly descriptions not only enhance detection accuracy but also improve the interpretability of the detection results. Specifically, we calculate the similarity between image features and each detailed anomaly description's text features. By examining the content of text descriptions with high similarity, we determine the specific anomaly category within the image, thereby gaining a deeper understanding of the model's decision-making process.

\subsubsection{Runtime Prompt Filtering}

\begin{figure}[]
  \centering
  \includegraphics[width=\linewidth]{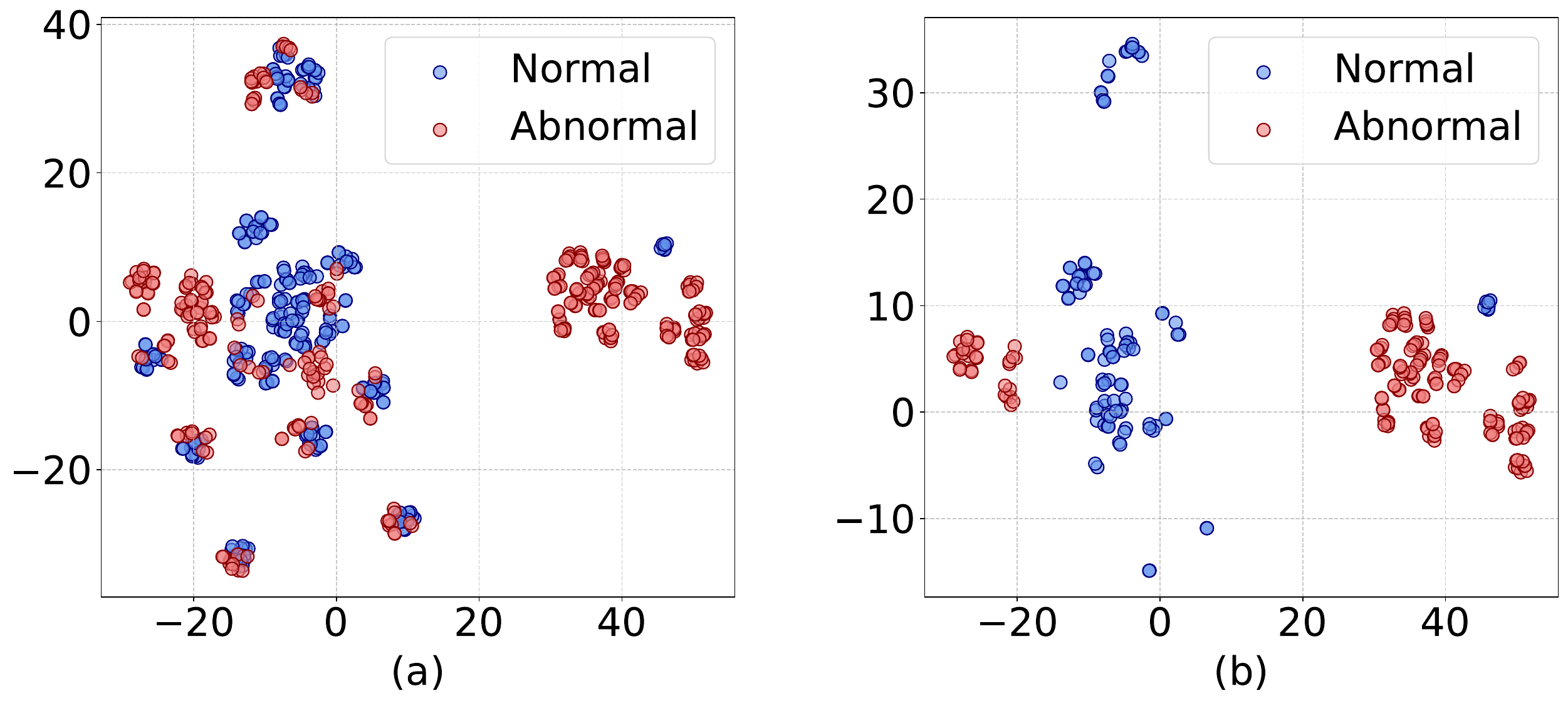} 
  \caption{t-SNE visualization results of textual feature vectors before and after Runtime Prompt Filtering implementation. Panel (a) represents the distribution prior to Runtime Prompt Filtering application, whereas panel (b) demonstrates the distribution subsequent to Runtime Prompt Filtering.}
  \label{fig:rtf}
\end{figure}

Ideally, anomaly detection methods based on image-text multimodal models should associate normal images with normal texts and anomalous images with anomalous texts, such that the similarity between normal image features and normal text features exceeds that between normal images and anomalous texts, and vice versa. However, in practice, we observe overlapping distances between normal and anomalous text features, a phenomenon termed cross-semantic ambiguity~\cite{zhu2024llms}, which hinders anomaly detection. Specifically, for the set of normal text features \( T_{n}^{origin} = \{T_{n,1}, T_{n,2}, \dots\} \) and anomalous text features \( T_a^{origin} = \{T_{a,1}, T_{a,2}, \dots\} \), we first compute the cosine distances between the test image features and each feature in both sets, resulting in distance sets \( D_n = \{D_{n,1}, D_{n,2}, \dots\} \) and \( D_a = \{D_{a,1}, D_{a,2}, \dots\} \). Ideally, for an image, the distance sets \( D_n \) and \( D_a \) should be mutually exclusive or have minimal overlap, meaning that for normal images, \( D_n \) should be significantly smaller than \( D_a \), and for anomalous images, \( D_a \) should be significantly smaller than \( D_n \). However, due to the diversity of text descriptions, not all text descriptions are reflected in a single test image, resulting in some noisy descriptions that may negatively impact model performance, as shown in Fig~\ref{fig:rtf}(a).

To address this issue, we design a strategy to filter the overlapping prompts in \( D_n \) and \( D_a \). Specifically, we first  determine the overlapping interval as follws: \\
\begin{equation*}
\begin{aligned}
     D_c = [\max(\min(D_n), \min(D_a)) + \alpha,  \\
     \min(\max(D_n), \max(D_a)) - \alpha].
\end{aligned}
\end{equation*}
where $\alpha$ is the hyperparameter for controlling the filtering.

We then remove the prompts in \( D_n \) and \( D_a \) that fall within \( D_c \), thereby completing runtime prompt filtering and obtaining the filtered text feature sets \( T_n \) and \( T_a \), as shown in Fig~\ref{fig:rtf}(b).

Next, we compute the global anomaly score by calculating the similarity between the global image feature \( G \), obtained by passing the image through the CLIP image encoder and adapter, and the filtered text features \( T_n \) and \( T_a \):
\begin{equation}
    S_{global} = \text{softmax}(G \cdot [T_n, T_a]^T) + \max(M),
\end{equation}
where \( M \) denotes the anomaly map calculated in Section~\ref{sec:DefLoc}, and \( \max(\cdot) \) represents the maximum operation. The adapter has a bottleneck structure to align global image features and text features, consisting of two linear layers, one ReLU~\cite{glorot2011deep} layer, and one SiLU~\cite{elfwing2018sigmoid} layer. Specifically, given the input feature $\mathbf{x}$, the output of the adapter is calculated as:
\begin{equation}
    \text{Adapter}(\mathbf{x}) = \text{SiLU}(\mathbf{W}_2 (\text{ReLU}(\mathbf{W}_1 \mathbf{x} + \mathbf{b}_1)) + \mathbf{b}_2),
\end{equation}
where $\mathbf{W}_1, \mathbf{b}_1$ and $\mathbf{W}_2, \mathbf{b}_2$ denote the weights and biases of the first and second linear layers, respectively.

The FusDes module not only improves anomaly detection accuracy but also enhances the interpretability of the detection results. By evaluating the similarity between image features and each detailed description's text features, we can identify the text descriptions that best match the image, thereby determining the specific anomaly types present in the image.

\subsection{DefLoc}
\label{sec:DefLoc}

Existing anomaly detection methods locate anomalous patches by directly computing the similarity between each image patch's features and text features. However, an anomalous region often spans multiple patches, and different anomalous regions vary in position, shape, and size. Sometimes, it is necessary to consider the surrounding normal regions to determine whether a region is anomalous. To address these challenges, we design the DefLoc module, which utilizes initial localization via Grounding DINO, position-enhanced text prompts, and a Multi-scale Deformable Cross-modal Interaction (MDCI) module to accurately locate anomalous regions of varying sizes and shapes. Additionally, we introduce a few-shot anomaly detection branch that employs a position-enhanced patch matching approach to achieve more accurate few-shot anomaly detection and localization results.

\subsubsection{Initial Localization with Grounding DINO}

Previous multimodal pre-trained model-based anomaly detection methods typically treat patches from different image locations equally when computing similarity with text features. However, the object under inspection often occupies only a portion of the input image, with the remaining background requiring exclusion. Direct patch similarity computation may erroneously identify minor disturbances in the background as anomalous regions, leading to false detections. We utilize the detailed anomaly descriptions generated by FusDes and employ the Grounding DINO method for initial anomaly localization. Although Grounding DINO alone cannot precisely determine the exact anomaly locations, the resulting bounding boxes generally encompass the foreground objects. Therefore, we use Grounding DINO's localization results to restrict the anomaly regions, effectively avoiding false detections in the background and improving the accuracy of subsequent anomaly localization. Furthermore, since Grounding DINO may occasionally miss some anomalies, we do not adopt an all-or-nothing approach for regions outside the bounding boxes. Instead, we suppress the anomaly scores of regions outside all Grounding DINO bounding boxes by multiplying them by a hyperparameter \( \lambda \), thereby mitigating potential misses caused by Grounding DINO.

\subsubsection{Position-Enhanced Text Prompts}

After obtaining initial anomaly localization results with Grounding DINO, we incorporate the positional information of the bounding boxes into the text prompts to enhance position descriptions. Specifically, we categorize the anomaly positions within the image into nine regions: top-left, left, bottom-left, top, center, bottom, top-right, right, and bottom-right. Based on the center coordinates of the Grounding DINO bounding boxes, we determine which positions are likely to contain anomalies and add this positional information to the \textit{[pos]} field of the text prompts. Text prompts enriched with detailed anomaly descriptions and positional information better match the content of the image under inspection, facilitating the model to focus on specific regions during subsequent anomaly localization and thereby improving localization precision.

\subsubsection{Multi-scale Deformable Cross-modal Interaction}

To accurately localize anomalous regions of different shapes and sizes, our method does not directly compute the similarity between each image patch feature and text features. Instead, we design a Multi-scale Deformable Cross-modal Interaction (MDCI) module. The design of MDCI is inspired by WinCLIP's approach of selecting image subregions using sliding windows of different sizes, but it overcomes the computational overhead of inputting dozens of differently sized images into the Image Encoder simultaneously as in WinCLIP. Specifically, we design deformable convolution kernels of varying sizes and shapes to aggregate regions of the patch features extracted by the CLIP Image Encoder in parallel. We then compute the similarity between the aggregated features and the position-enhanced text features, performing text feature-guided multi-scale deformable convolution operations. This approach allows the MDCI module to handle anomalous regions of varying sizes and shapes simultaneously, significantly enhancing the model's ability to localize anomalies while maintaining high computational efficiency.

Specifically, we design \( n \) deformable convolution kernels of different sizes and shapes, denoted as \( D_j \), where \( j \) ranges from 1 to \( n \). For each stage \( i \), the patch features \( P_i \in \mathbb{R}^{H_iW_i\times C} \) extracted by image encoder, along with the position-enhanced text features \( [T_n, T_a] \in \mathbb{R}^{2 \times C} \), are processed as follows:
\begin{equation}
    M^n_i, M^a_i = \text{Up}(\text{Norm}(\sum_{j=1}^{n} \text{S}(D_j(P_i) \cdot [T_n, T_a]^T))),
\end{equation}
where \( \text{Up}(\cdot) \) denotes upsampling, \( \text{S}(\cdot) \) represents the Softmax operation, and \( \text{Norm}(\cdot) \) denotes the normalization operation. By summing and normalizing the maps from each stage, we obtain the final normal and anomaly maps:
\begin{equation}
    M^n = \text{Norm}\left(\sum_i M^n_i\right), \quad M^a = \text{Norm}\left(\sum_i M^a_i\right).
\end{equation}

The anomaly localization result obtained through this vision-language feature comparison method is expressed as:
\begin{equation}
    M^{vl} = \frac{M^a + (1 - M^n)}{2}.
\end{equation}

\subsubsection{Few-shot Anomaly Detection Branch} 

In the newly added few-shot anomaly detection branch, we design a position-enhanced patch matching method that utilizes the initial localization information obtained from DefLoc to constrain the patch matching regions. Specifically, in the few-shot anomaly detection branch, we first use the same CLIP image encoder to extract patch-level features from known normal image samples and store each stage's features in the corresponding memory bank. Then, for each stage \( i \) of the query image's patch features \( P_i \), we calculate the cosine distance between \( P_i \) and all normal patch features in the corresponding memory bank, selecting the minimum cosine distance as the anomaly score for that patch at that stage:
\begin{equation}
    M_i^{few} = \min(\text{cos\_distance}(P_i, \text{Mem}_i)),
\end{equation}
where \( \text{Mem}_i \) represents the memory bank for stage \( i \). By summing and normalizing the anomaly scores across all stages and suppressing the scores of regions outside the Grounding DINO bounding boxes, we obtain the few-shot anomaly localization result:
\begin{equation}
    M^{few} = \text{Norm}\left(\sum_i M^{few}_i\right).
\end{equation}

Combining the few-shot anomaly localization results with the image-text anomaly localization results, we obtain the final anomaly localization result:
\begin{equation}
    M = G_{\sigma}\left(\frac{M^{vl} + M^{few}}{2}\right),
\end{equation}
where \( G_{\sigma} \) is a Gaussian filter, and \( \sigma \) is a hyperparameter controlling the degree of smoothing, set to 4 in our experiments.

\subsection{Loss Functions}
To learn the content of adaptive text templates and the parameters in MDCI and adapter, we choose different loss functions for training from the perspectives of global anomaly detection and local anomaly localization. We choose the cross-entropy loss as our global loss for global anomaly detection, as it is a commonly used binary classification loss function. The global loss is computed as follows:

\begin{equation}
    L_{global} = L_{ce}(S_{global}, Label),
\end{equation}
where \(S_{\text{global}}\) is the global anomaly score computed in Sec.~\ref{sec:FusDes}, and \textit{Label} indicates whether the image is anomalous or not. For local anomaly localization, we employ Focal loss~\cite{lin2017focal} and Dice loss~\cite{milletari2016v} to optimize the anomaly map. Focal Loss and Dice Loss are common loss functions used in semantic segmentation tasks. Specifically, Focal Loss is particularly effective in addressing class imbalance issues, making it well-suited for anomaly localization tasks where the proportion of anomaly regions is relatively small. Focal loss can be calculated by Eq.~(\ref{eq:focal}):
\begin{equation}
    L_{f} = -\frac{1}{n}\sum_{i=1}^{n}{(1-p_i)^{\gamma}log(p_i)},\label{eq:focal}
\end{equation}
where $n = H \times W$ represents the total number of pixels, $p_i$ is the predicted probability of the positive classes and $\gamma$ is a tunable parameter for adjusting the weight of hard-to-classify samples. In our implementation, we set $\gamma$ to 2.

Dice loss can be calculated by Eq.~(\ref{eq:dice}):
\begin{equation}
    L_{d} = -\frac{\sum_{i=1}^{n}{y_i\hat{y}_i}}{\sum_{i=1}^{n}{y_i^2}+\sum_{i=1}^{n}{\hat{y}^2_i}},\label{eq:dice}
\end{equation}
where $n = H \times W$, $y_i$ is the output of decoder and $\hat{y}_i$ is the ground truth value. 

Our local loss can be calculated by Eq.~(\ref{eq:local}):
\begin{equation}
    L_{local} = L_{f}(M^a, G) + L_{d}(M^a, G) + L_{d}(M^n, 1 - G),\label{eq:local}
\end{equation}
where \(G\) denotes the ground truth.

\section{Experiments}
\subsection{Datasets}
We conduct experiments primarily on the MVTec-AD~\cite{bergmann2019mvtec} and VisA~\cite{zou2022spot} datasets. MVTec-AD is a comprehensive industrial anomaly detection dataset, comprising 5,354 images from 15 different categories, including 10 object categories and 5 texture categories, with resolutions ranging from $700\times 700$ to $1,024\times 1,024$ pixels. In comparison, the VisA dataset presents greater challenges, containing 10,821 images from 12 different categories, with resolutions of $1,500\times 1,000$ pixels. Consistent with APRIL-GAN~\cite{chen2023april} and AnomalyCLIP~\cite{zhou2023anomalyclip}, we perform supervised training on one dataset and conduct zero-shot or few-shot testing on the other dataset.

To evaluate the generalization ability of the model, we also conduct experiments on medical datasets: BrainMRI~\cite{bao2024bmad} and RESC~\cite{hu2019automated}. The BrainMRI dataset is based on the BraTS2021~\cite{baid2021rsna} dataset, one of the latest large-scale brain tumor segmentation datasets. The BrainMRI dataset consists of 2D slices derived from BraTS2021, with each slice image measuring 240$\times$240 pixels. The training set includes 7,500 normal samples, and the test set contains 3,715 samples, both normal and anomalous, with pixel-level anomaly annotations. The Retinal Edema Segmentation Challenge (RESC) dataset is a retinal OCT dataset containing 4,297 normal images for training and 1,805 test images, both normal and anomalous. The image resolution is 512$\times$1,024, and the dataset provides pixel-level anomaly annotations.

\subsection{Evaluation Metrics}
Consistent with existing zero-shot and few-shot anomaly detection methods~\cite{jiang2024fabgpt, you2022unified, gu2024anomalygpt}, we employ the Area Under the Receiver Operating Characteristic (AUC) as the evaluation metric. Image-level AUC is utilized to assess the performance of anomaly detection, while pixel-level AUC evaluates the performance of anomaly localization.

\subsection{Implementation Details}
We employ the CLIP-L/14@336px model as the backbone, freezing the parameters of both the CLIP text encoder and image encoder. Training is conducted on either the MVTec-AD or VisA dataset, while zero-shot and few-shot testing is performed on the other dataset. For the experiments on medical datasets, we utilize the model weights trained on the MVTec-AD dataset to evaluate generalization capability. For intermediate layer patch features, we utilize the features from the 6th, 12th, 18th, and 24th layers of the CLIP image encoder. Starting from the 7th layer, we simultaneously leverage the outputs of QKV Attention and V-V Attention, where the output of QKV Attention is aligned with text features through a simple linear layer, and the output of V-V Attention is input into the MDCI module for multi-scale and multi-shape deep interaction with text features. During training, input images are resized to a resolution of $518 \times 518$ and the model parameters are optimized for 15 epochs using the AdamW~\cite{loshchilov2017decoupled} optimizer. The learning rate for the learnable text vectors is set to 1e-3, while the learning rate for the MDCI module is set to 1e-4. Subsequently, we train the adapter for 5 epochs with a learning rate of 1e-5. Additionally, due to the varying number of fine-grained anomaly descriptions across object categories, training is conducted with a batch size of 1. Following previous methods~\cite{you2022unified, zhou2023anomalyclip}, a Gaussian filter with $\sigma=4$ is applied during testing to obtain smoother anomaly score maps. We use the GPT-4o model in our paper and the prompt that we use is: \textit{``Based on your knowledge, what anomalies might occur on [class name]?"}. All experiments are conducted on a NVIDIA RTX A6000 GPU with 48GB memory.

\begin{table}[]
\centering
\caption{Comparison results between FiLo++ and other ZSAD methods. \textit{Img-AUC} and \textit{Px-AUC} in table represent image-level AUC and pixel-level AUC. The best-performing method is in \textbf{bold}.}
\label{tab:zero-results}
\begin{tabular}{@{}ccccc@{}}
\toprule
\multirow{2}{*}{Method}                                      & \multicolumn{2}{c}{MVTec-AD}      & \multicolumn{2}{c}{VisA}  \\ \cmidrule(l){2-5}
                                                                                                   & Img-AUC     & Px-AUC     & Img-AUC     & Px-AUC     \\ \midrule
CLIP~\cite{radford2021learning}                                    & 74.1          & 38.4   & 66.4          & 46.6        \\
CLIP-AC~\cite{radford2021learning}                                  & 71.5          & 38.2    & 65.0          & 47.8      \\
WinCLIP~\cite{jeong2023winclip}                                         & 91.8          & 85.1   & 78.1          & 79.6       \\
APRIL-GAN~\cite{chen2023april}                                          & 86.1          & 87.6  & 78.0          & 94.2        \\
AnomalyCLIP~\cite{zhou2023anomalyclip}                                & 91.5          & 91.1  & 82.1          & 95.5        \\
FiLo~\cite{gu2024filo}                                                                                & 91.2          & 92.3      & 83.9          & 95.9    \\
\textbf{FiLo++~(ours)}                                                & \textbf{92.1} & \textbf{92.8} & \textbf{84.5} & \textbf{96.2} \\ \bottomrule
\end{tabular}
\end{table}

\begin{table}[]
\centering
\caption{Comparison results between FiLo++ and other FSAD methods. \textit{Img-AUC} and \textit{Px-AUC} in table represent image-level AUC and pixel-level AUC. The best-performing method is in \textbf{bold}.}
\label{tab:few-results}
\begin{tabular}{@{}cccccc@{}}
\toprule
\multirow{2}{*}{Setup}  & \multirow{2}{*}{Method} & \multicolumn{2}{c}{MVTec-AD}  & \multicolumn{2}{c}{VisA}      \\ \cmidrule(l){3-6} 
                        &                         & Img-AUC     & Px-AUC     & Img-AUC     & Px-AUC     \\ \midrule
\multirow{5}{*}{1-shot} & SPADE                   & 81.0          & 91.2          & 79.5          & 95.6          \\
                        & PatchCore               & 83.4          & 92.0          & 79.9          & 95.4          \\
                        & WinCLIP                 & 93.1          & 95.2          & 83.8          & 96.4          \\
                        & AnomalyGPT              & 94.1          & 95.3          & 87.4          & 96.2          \\ \cmidrule(l){2-6} 
                        & \textbf{FiLo++ (ours)}  & \textbf{95.0} & \textbf{96.2} & \textbf{88.3} & \textbf{97.3} \\ \midrule
\multirow{5}{*}{2-shot} & SPADE                   & 82.9          & 92.0          & 80.7          & 96.2          \\
                        & PatchCore               & 86.3          & 93.3          & 81.6          & 96.1          \\
                        & WinCLIP                 & 94.4          & 96.0          & 84.6          & 96.8          \\
                        & AnomalyGPT              & 95.5          & 95.6          & \textbf{88.6} & 96.4          \\ \cmidrule(l){2-6} 
                        & \textbf{FiLo++ (ours)}  & \textbf{95.8} & \textbf{96.5} & \textbf{88.6} & \textbf{97.5} \\ \midrule
\multirow{5}{*}{4-shot} & SPADE                   & 84.8          & 92.7          & 81.7          & 96.6          \\
                        & PatchCore               & 88.8          & 94.3          & 85.3          & 96.8          \\
                        & WinCLIP                 & 95.2          & 96.2          & 87.3          & 97.2          \\
                        & AnomalyGPT              & \textbf{96.3} & 96.2          & \textbf{90.6} & 96.7          \\ \cmidrule(l){2-6} 
                        & \textbf{FiLo++ (ours)}  & \textbf{96.3} & \textbf{96.6} & 89.8          & \textbf{97.9} \\ \bottomrule
\end{tabular}
\end{table}

\begin{table}[]
\centering

\caption{Few-shot anomaly detection results on medical datasets (BrainMRI and RESC). \textit{Img-AUC} and \textit{Px-AUC} represent image-level and pixel-level AUC. The best-performing method is in \textbf{bold}.}
\label{tab:medical}
\begin{tabular}{@{}cccccc@{}}
\toprule
\multirow{2}{*}{Setup}  & \multirow{2}{*}{Method} & \multicolumn{2}{c}{BrainMRI}  & \multicolumn{2}{c}{RESC}      \\ \cmidrule(l){3-6} 
                        &                         & Img-AUC       & Px-AUC        & Img-AUC       & Px-AUC        \\ \midrule
\multirow{5}{*}{1-shot} & PatchCore               & 73.2          & 96.0          & 56.3          & 78.2          \\
                        & AnomalyGPT              & 73.1          & 96.0          & 82.4          & 94.0          \\
                        & WinCLIP                 & 55.4          & 86.6          & 72.9          & 87.9          \\
                        & MedCLIP                 & 69.7          & 91.7          & 66.9          & 91.5          \\ \cmidrule(l){2-6} 
                        & \textbf{FiLo++ (ours)}  & \textbf{86.6} & \textbf{96.3} & \textbf{86.8} & \textbf{94.7} \\ \midrule
\multirow{5}{*}{2-shot} & PatchCore               & 60.4          & 94.2          & 69.1          & 81.8          \\
                        & AnomalyGPT              & 73.4          & 95.3          & 86.2          & \textbf{94.7} \\
                        & WinCLIP                 & 55.1          & 86.5          & 74.1          & 87.7          \\
                        & MedCLIP                 & 69.8          & 91.8          & 76.2          & 93.1          \\ \cmidrule(l){2-6} 
                        & \textbf{FiLo++ (ours)}  & \textbf{86.7} & \textbf{96.3} & \textbf{87.4} & \textbf{94.7} \\ \midrule
\multirow{5}{*}{4-shot} & PatchCore               & 75.6          & 91.4          & 72.1          & 88.5          \\
                        & AnomalyGPT              & 74.6          & 95.4          & 84.9          & \textbf{95.2} \\
                        & WinCLIP                 & 58.3          & 86.6          & 71.3          & 87.0          \\
                        & MedCLIP                 & 73.7          & 91.5          & 83.0          & 91.2          \\ \cmidrule(l){2-6} 
                        & \textbf{FiLo++ (ours)}  & \textbf{86.9} & \textbf{96.4} & \textbf{87.6} & 95.1          \\ \bottomrule
\end{tabular}

\end{table}

\begin{table}[]
\centering
\caption{Ablation Results for Anomaly Descriptions. \textit{GS} stands for generic State, \textit{FGD} for Fine-Grained Description, \textit{LT} for Learnable Template, and \textit{RTPF} for Runtime Prompt Filtering. \textit{Img-AUC} and \textit{Px-AUC} represent image-level and pixel-level AUC, respectively. The best-performing result is in \textbf{bold}.}
\label{tab:ab-fusdes}
\begin{tabular}{@{}ccccc@{}}
\toprule
\multirow{2}{*}{Setup} & \multicolumn{2}{c}{MVTec-AD} & \multicolumn{2}{c}{VisA} \\ \cmidrule(l){2-5} 
                       & Img-AUC       & Px-AUC       & Img-AUC     & Px-AUC     \\ \midrule
CLIP baseline          & 71.5          & 38.2         & 65.0        & 47.8       \\
+ GS                   & 79.9          & 83.5         & 65.4        & 83.9       \\
+ FGD                  & 80.8          & 83.8         & 71.2        & 85.5       \\
+ LT                   & 85.8          & 85.1         & 78.1        & 93.2       \\
+ \textbf{RTPF}                 & \textbf{86.2}          & \textbf{85.3}         & \textbf{78.5}        & \textbf{93.5}       \\ \bottomrule
\end{tabular}
\end{table}

\begin{table*}[]
\centering
\caption{The results of ablation experiments for each proposed modules in DefLoc. The best-performing result is in \textbf{bold}.}
\label{tab:defloc}
\begin{tabular}{@{}ccccccccc@{}}
\toprule
\multirow{2}{*}{Grounding} &
  \multirow{2}{*}{Position Enhancement} &
  \multicolumn{3}{c}{MDCI} &
  \multicolumn{2}{c}{MVTec-AD} &
  \multicolumn{2}{c}{VisA} \\ \cmidrule(l){6-9} 
             &              & Multi-shape  & Multi-scale & Deformable  & Image-AUC & Pixel-AUC & Image-AUC & Pixel-AUC \\ \midrule
             &              &              &      &              & 86.2      & 85.3 & 78.5      & 93.5     \\
$\checkmark$ &              &              &      &              & 86.2      & 85.7  & 78.5      & 93.9    \\
$\checkmark$ & $\checkmark$ &              &     &               & 86.5      & 85.9  & 78.6      & 94.1    \\
$\checkmark$ & $\checkmark$ & $\checkmark$ &     &               & 86.8      & 89.6  & 79.4      & 95.6    \\
$\checkmark$ & $\checkmark$ &              & $\checkmark$ &       & 89.3      & 91.7  & 81.2      & 95.9    \\
$\checkmark$ &
  $\checkmark$ &
  $\checkmark$ &
  $\checkmark$ &
    &
  91.4 &
  92.3
  &
  84.1 &
  95.9
  \\ 
$\checkmark$ &
  $\checkmark$ &
  $\checkmark$ &
  $\checkmark$ &
  $\checkmark$  &
  \textbf{92.1} &
  \textbf{92.8} &
  \textbf{84.5} &
  \textbf{96.2}
  \\ 
  
  \bottomrule
\end{tabular}

\end{table*}

\subsection{Zero-shot Results}
To demonstrate the effectiveness of our proposed FiLo++, we compare it against several existing zero-shot anomaly detection methods, including CLIP~\cite{radford2021learning}, CLIP-AC~\cite{radford2021learning}, WinCLIP~\cite{jeong2023winclip}, APRIL-GAN~\cite{chen2023april}, AnomalyCLIP~\cite{zhou2023anomalyclip} and FiLo~\cite{gu2024filo}. Following the methodology of AnomalyCLIP~\cite{zhou2023anomalyclip}, we conduct experiments with CLIP using straightforward text prompts such as \textit{``A photo of a normal [class]."} and \textit{``A photo of an anomalous [class]."} Additionally, for CLIP-AC, we incorporate a variety of text prompt templates recommended for the ImageNet dataset to enhance performance. The results for WinCLIP~\cite{jeong2023winclip}, APRIL-GAN~\cite{chen2023april}, and AnomalyCLIP~\cite{zhou2023anomalyclip} are directly adopted from their respective publications.

Table~\ref{tab:zero-results} presents a performance comparison between FiLo++ and other zero-shot anomaly detection methods. The results indicate that FiLo++ outperforms existing ZSAD approaches in both anomaly detection and localization, thereby demonstrating the effectiveness of the FusDes and DefLoc modules we have designed.

\subsection{Few-shot Results on Industrial Datasets}
In order to further validate the effectiveness of FiLo++ in few-shot anomaly detection, we compared its performance against several state-of-the-art few-shot anomaly detection methods, including SPADE~\cite{cohen2020sub}, PatchCore~\cite{roth2022towards}, WinCLIP~\cite{jeong2023winclip}, and AnomalyGPT~\cite{gu2024anomalygpt}. In this setting, unlike the zero-shot scenario, a small number of normal samples from the test classes are provided for reference during testing. We conducted experiments under 1-shot, 2-shot, and 4-shot configurations respectively.

Table~\ref{tab:few-results} presents a performance comparison between our FiLo++ and other few-shot anomaly detection methods. The results reveal that FiLo++ achieves a marked performance improvement over conventional patch-matching approaches under few-shot settings, particularly in the 1-shot scenario. This highlights the effectiveness of FiLo++'s vision-language matching and position-enhanced patch matching approach in few-shot anomaly detection.

\subsection{Generalization on Medical Datasets}

    To evaluate the generalization capability and robustness of FiLo++ across different domains, we extended our experiments to medical anomaly detection datasets: BrainMRI~\cite{bao2024bmad} and RESC~\cite{hu2019automated}. We compared FiLo++ against general anomaly detection methods PatchCore~\cite{roth2022towards}, WinCLIP~\cite{jeong2023winclip}, AnomalyGPT~\cite{gu2024anomalygpt} and a domain-specific method, MedCLIP~\cite{wang2022medclip}.

    As shown in Table~\ref{tab:medical}, FiLo++ demonstrates superior performance in the medical domain. Notably, on the BrainMRI dataset, FiLo++ achieves an Image-level AUC of 86.6\% in the 1-shot setting, significantly outperforming the second-best method, AnomalyGPT~\cite{gu2024anomalygpt}, by a margin of 13.5\%, and surpassing the medical-specific MedCLIP~\cite{wang2022medclip} by 16.9\%. Similarly, on the RESC dataset, FiLo++ consistently maintains the highest Image-level AUC across all shot settings. 

    It is worth noting that while generic methods like WinCLIP~\cite{jeong2023winclip} struggle with the domain shift in medical imagery, and even domain-specific methods like MedCLIP~\cite{wang2022medclip} show limitations in few-shot scenarios, FiLo++ effectively adapts to these challenges. This success can be attributed to the FusDes module's ability to generate precise, domain-aware descriptions for medical pathologies and the DefLoc module's capacity to localize subtle lesions, proving that our position-enhanced patch matching strategy is highly effective even in specialized non-industrial domains.

\begin{table}[]
\centering
\caption{Ablation Results for Few-shot branch. \textit{PM} stands for original patch matching, \textit{GDINO} for Grounding DINO, and \textit{VL} for vision-language feature matching. \textit{Img-AUC} and \textit{Px-AUC} represent image-level and pixel-level AUC, respectively. The best-performing method is in \textbf{bold}.}
\label{tab:ab-few}
\begin{tabular}{@{}cccccc@{}}
\toprule
\multirow{2}{*}{Setup}  & \multirow{2}{*}{Method} & \multicolumn{2}{c}{MVTec-AD}  & \multicolumn{2}{c}{VisA}      \\ \cmidrule(l){3-6} 
                        &                         & Img-AUC       & Px-AUC        & Img-AUC       & Px-AUC        \\ \midrule
\multirow{3}{*}{1-shot} & PM          & 94.9          & 94.8          & 80.6          & \textbf{97.3} \\
                        & + GDINO        & \textbf{95.0} & \textbf{96.2} & 81.7          & \textbf{97.3} \\
                        & + VL       & \textbf{95.0} & \textbf{96.2} & \textbf{88.3} & \textbf{97.3} \\ \midrule
\multirow{3}{*}{2-shot} & PM          & \textbf{95.8} & 95.3          & 80.8          & \textbf{97.5} \\
                        & + GDINO        & \textbf{95.8} & 96.5          & 82.0          & \textbf{97.5} \\
                        & + VL       & \textbf{95.8} & \textbf{96.5} & \textbf{88.6} & \textbf{97.5} \\ \midrule
\multirow{3}{*}{4-shot} & PM          & \textbf{96.3} & 95.6          & 85.6          & \textbf{97.9} \\
                        & + GDINO        & \textbf{96.3} & 96.6          & 87.3          & \textbf{97.9} \\
                        & + VL       & \textbf{96.3} & \textbf{96.6} & \textbf{89.8} & \textbf{97.9} \\ \bottomrule
\end{tabular}
\end{table}

\begin{table}[]
\caption{Comparison of different learning methods for learnable vectors and whether to use class name. \textit{Img-AUC} and \textit{Px-AUC} represent image-level and pixel-level AUC, respectively. The best-performing method is in \textbf{bold}.}
\label{tab:coop}
\centering
\begin{tabular}{@{}cccccc@{}}
\toprule
\multirow{2}{*}{\begin{tabular}[c]{@{}c@{}}Learning\\ method\end{tabular}} & \multirow{2}{*}{\begin{tabular}[c]{@{}c@{}}CLS\\ name\end{tabular}} & \multicolumn{2}{c}{MVTec-AD}  & \multicolumn{2}{c}{VisA}      \\ \cmidrule(l){3-6} 
                                                                           &                           & Img-AUC       & Px-AUC        & Img-AUC       & Px-AUC        \\ \midrule
CoOp                                                                       &                           & 90.1          & 88.8          & 82.0            & 95.5          \\
CoOp                                                                       & $\checkmark$              & 89.9          & 90.4          & 81.2          & 95.4          \\
CoCoOp                                                                     &                           & 91.5          & 90.8          & 82.7          & 95.7          \\
\textbf{CoCoOp}                                                            & $\checkmark$              & \textbf{92.1} & \textbf{92.8} & \textbf{84.5} & \textbf{96.2} \\ \bottomrule
\end{tabular}

\end{table}

\begin{figure}[t]
  \centering
  \includegraphics[width=0.9\linewidth]{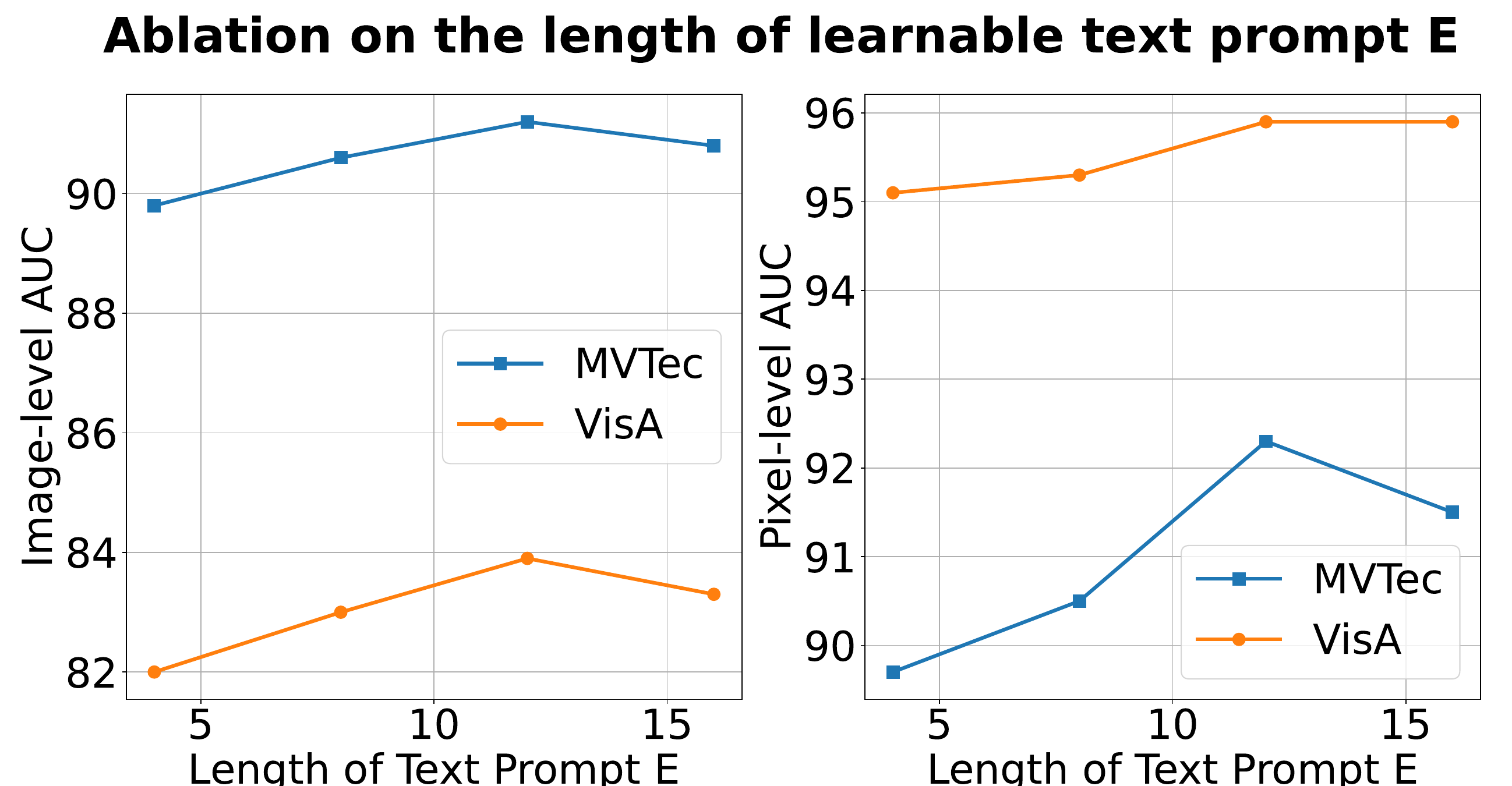} 
  \caption{Comparison of FiLo++ on MVTec and VisA datasets with different numbers of learnable vectors.}
  \label{fig:num_vec}
\end{figure}

\begin{figure}[t]
  \centering
  \includegraphics[width=0.9\linewidth]{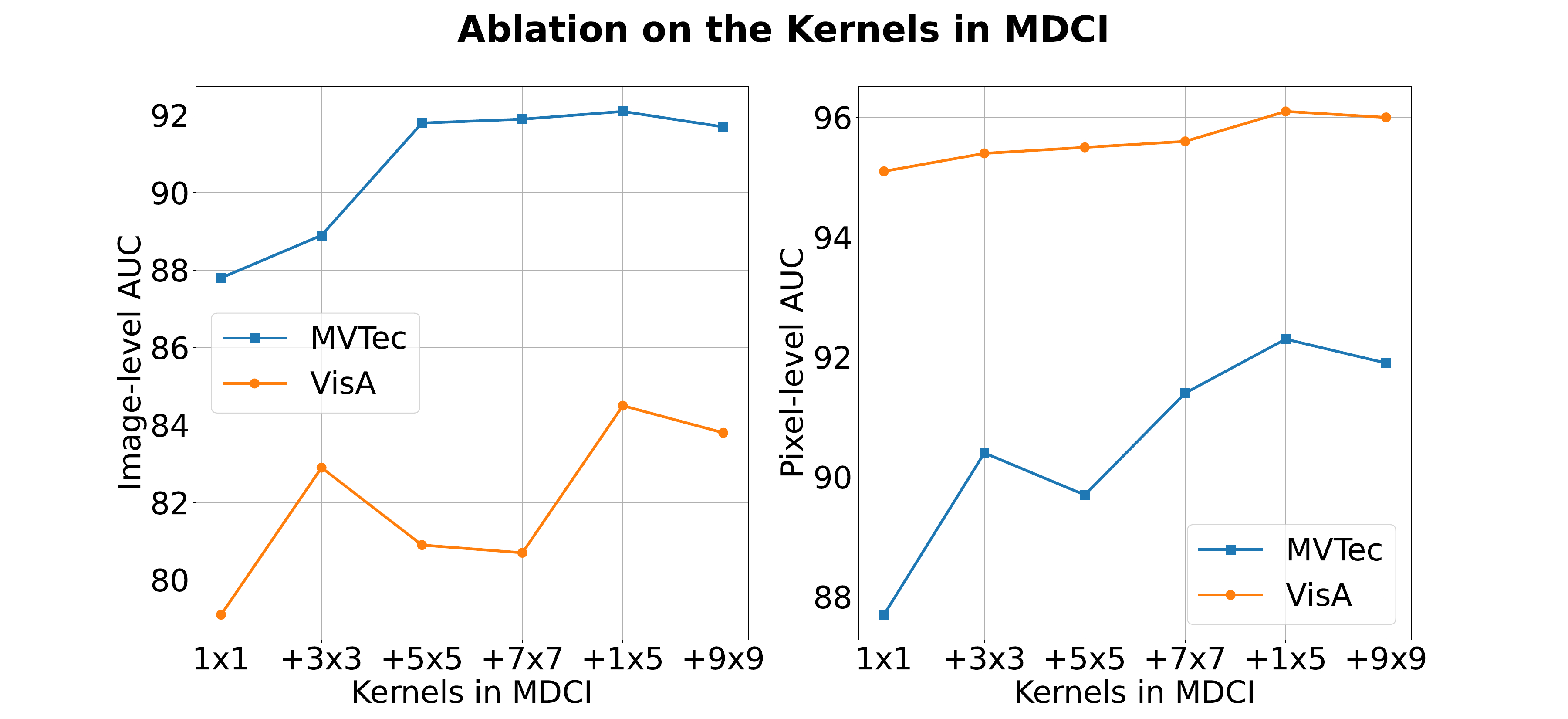} 
  \caption{Comparison of FiLo++ on MVTec and VisA datasets with different convolution kernels.}
  \label{fig:mdci_kernel}
\end{figure}

\begin{figure}[t]
  \centering
  
  \includegraphics[width=0.95\linewidth]{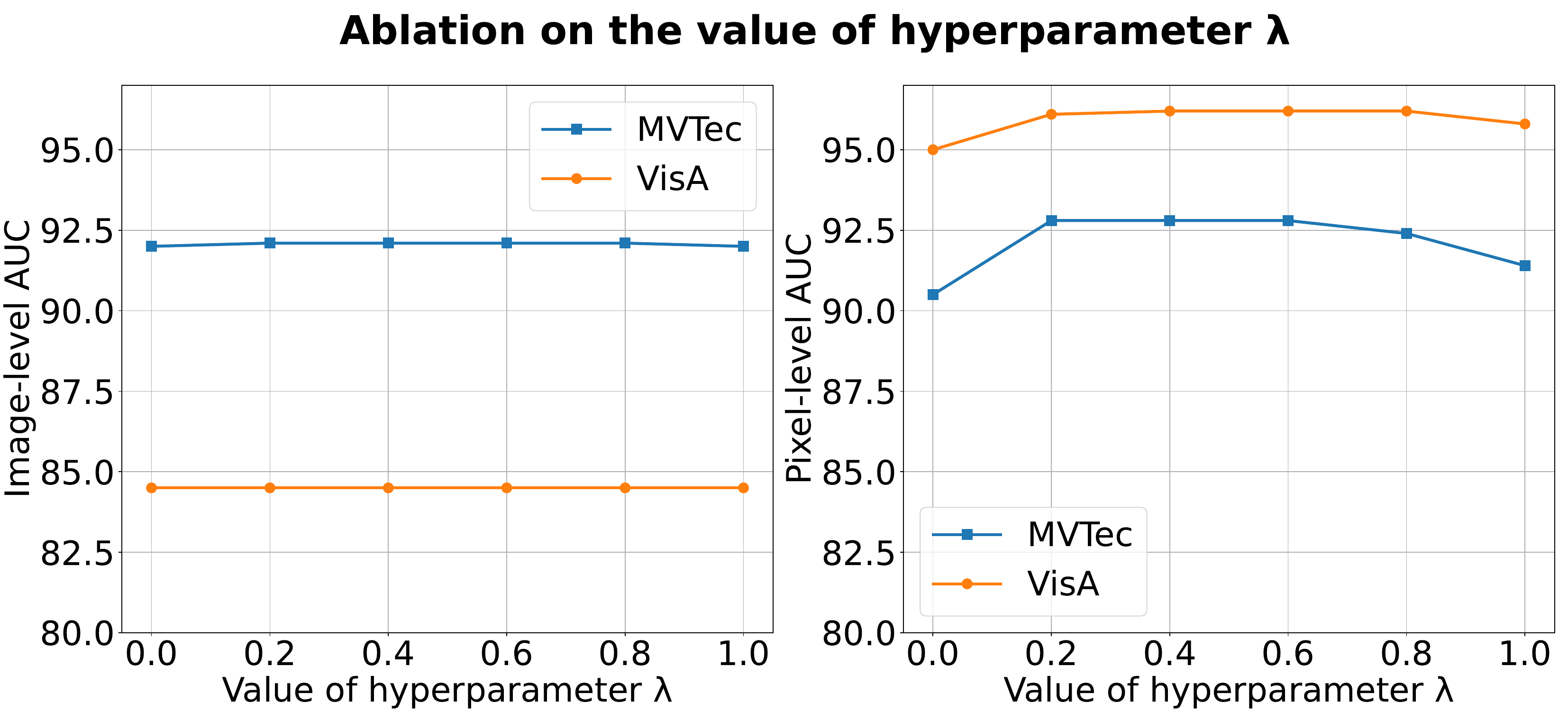} 
  \caption{Ablation on the value of hyperparameter $\lambda$ on MVTec-AD and VisA datasets. The left chart shows Image-level AUC, and the right chart shows Pixel-level AUC.}
  
  \label{fig:lambda}
\end{figure}

\begin{figure}[t]
  \centering
 
  \includegraphics[width=0.95\linewidth]{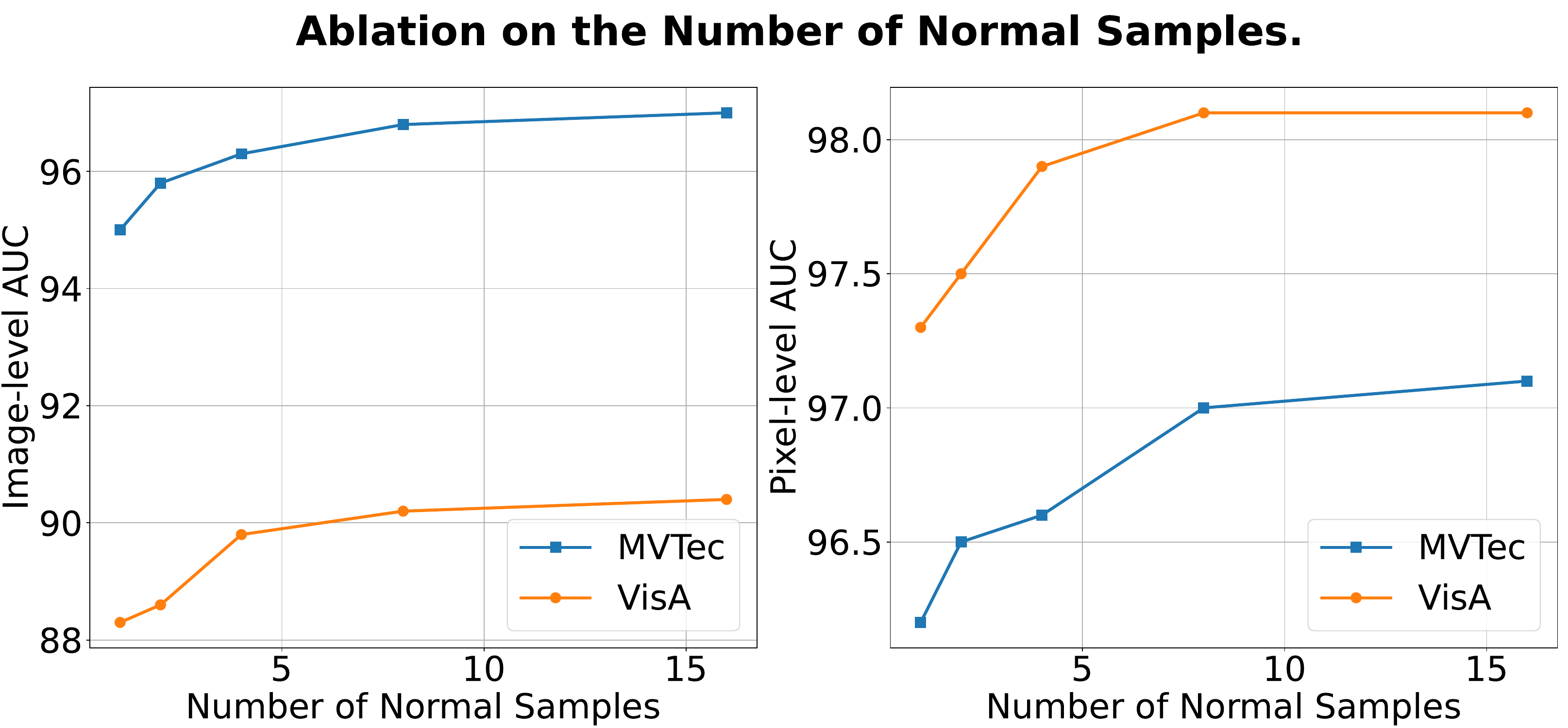} 
  \caption{ Ablation study on the number of normal reference samples ($K$) for few-shot anomaly detection. The results demonstrate the performance trend on MVTec-AD and VisA datasets as $K$ increases from 1 to 16}.
  
  \label{fig:num_k}
\end{figure}

\subsection{Ablation Study}
In order to validate the effectiveness of each proposed module, we conducted extensive ablation experiments on the MVTec AD and VisA datasets. These experiments examined various components in both FusDes and DefLoc, the usage of vision-language features and location enhancement in few-shot anomaly detection, the implementation strategies for learnable text vectors, the application of V-V Attention, and the impact of different convolutional kernels in the MDCI module.

\subsubsection{FusDes Module}
In Table~\ref{tab:ab-fusdes}, We conduct ablation experiments on various parts of the FusDes module. We begin with the baseline model CLIP-AC, utilizing simple two-category texts \textit{``A photo of a normal [class]”} and \textit{``A photo of an anomalous [class]”}. Building upon this foundation, we progressively incorporate generic state descriptions, fine-grained anomaly descriptions generated by large language models, learnable prompt templates, and runtime prompt filtering strategies. The experimental results demonstrate that more detailed and customized textual descriptions significantly enhance the performance of anomaly detection methods based on vision-language feature matching.

\subsubsection{DefLoc Module}
In Table~\ref{tab:defloc}, we further conduct ablation experiments on the modules within DefLoc. Experimental results demonstrate that both Grounding DINO and Position Enhancement contribute to the improvement of pixel-level AUC. Additionally, the MDCI module integrates multi-scale and deformable functionalities, effectively detecting anomalies of various sizes and shapes, thereby enhancing both detection and localization performance.

\subsubsection{Few-shot branch}
Subsequently, in Table~\ref{tab:ab-few}, we assess the performance impact of Grounding DINO and vision-language feature comparison within FiLo++ on few-shot anomaly detection. The results indicate that both components of FiLo++ provide performance enhancements to the original patch-matching-based few-shot anomaly detection methods.

\begin{table}[t]
\centering
\caption{Comparison of results of different processing methods for the output results of QKV and VV Attention.}
\label{tab:qkvvv}
\begin{tabular}{@{}cccccc@{}}
\toprule
\multirow{2}{*}{QKV attn} & \multirow{2}{*}{VV attn} & \multicolumn{2}{c}{MVTec-AD}  & \multicolumn{2}{c}{VisA}      \\ \cmidrule(l){3-6} 
                     &                     & Img-AUC       & Px-AUC        & Img-AUC       & Px-AUC        \\ \midrule
\textbf{Linear}      & \textbf{MDCI}       & \textbf{92.1} & \textbf{92.8} & \textbf{84.5} & \textbf{96.2} \\
Linear               & Linear              & 90.0            & 90.2          & 82.7          & 95.4          \\
MDCI                 & MDCI                & 90.7          & 83.3          & 83.4          & 95.6          \\
MDCI                 & Linear              & 90.5          & 86.4          & 83.2          & 95.7          \\ \bottomrule
\end{tabular}
\end{table}

\begin{table}[t]
\centering
\caption{Comparison of different LLMs used to generate anomaly descriptions. The best-performing method is in \textbf{bold}.}
\label{tab:llm}
\begin{tabular}{@{}ccccc@{}}
\toprule
\multirow{2}{*}{LLM} & \multicolumn{2}{c}{VisA} & \multicolumn{2}{c}{MVTec-AD} \\ \cmidrule(l){2-5} 
                     & Image-AUC   & Pixel-AUC  & Image-AUC     & Pixel-AUC    \\ \midrule
\textbf{GPT-4o}                & \textbf{84.5}        & \textbf{96.2}       & \textbf{92.1}         & \textbf{92.8}         \\
GPT-3.5              & 84.3        & 96.2       & 91.8          & 92.6         \\
Llama                & 84.4          & 96.0       & 91.9            & 92.5         \\ \bottomrule \\[-0.6cm]
\end{tabular}
\end{table}

\begin{table}[t]
\centering

\caption{Comparison of inference time and GPU memory usage between FiLo++ with standard and deformable convolution.}
\label{tab:efficiency}
\begin{tabular}{@{}ccc@{}}
\toprule
                         & Inference Time & GPU Memory Usage \\ \midrule
FiLo++ w/ Standard Conv.   & 312 ms          & 7,526 MB          \\
FiLo++ w/ Deformable Conv. & 344 ms          & 7,783 MB          \\ \bottomrule
\end{tabular}

\end{table}

\begin{figure*}[t]
  \centering
  \includegraphics[width=0.9\textwidth]{images/visualization_tcsvt.pdf} 
  \caption{Qualitative comparison of anomaly localization results on MVTec-AD and VisA datasets. We compare the anomaly maps produced by PatchCore (1-shot), AnomalyGPT (1-shot), and our FiLo++ (both 0-shot and 1-shot). FiLo++ exhibits more precise localization and stronger background suppression compared to other state-of-the-art few-shot methods.}
  \label{fig:visual}
\end{figure*}

\begin{figure*}[]
  \centering
  
  \includegraphics[width=0.9\textwidth]{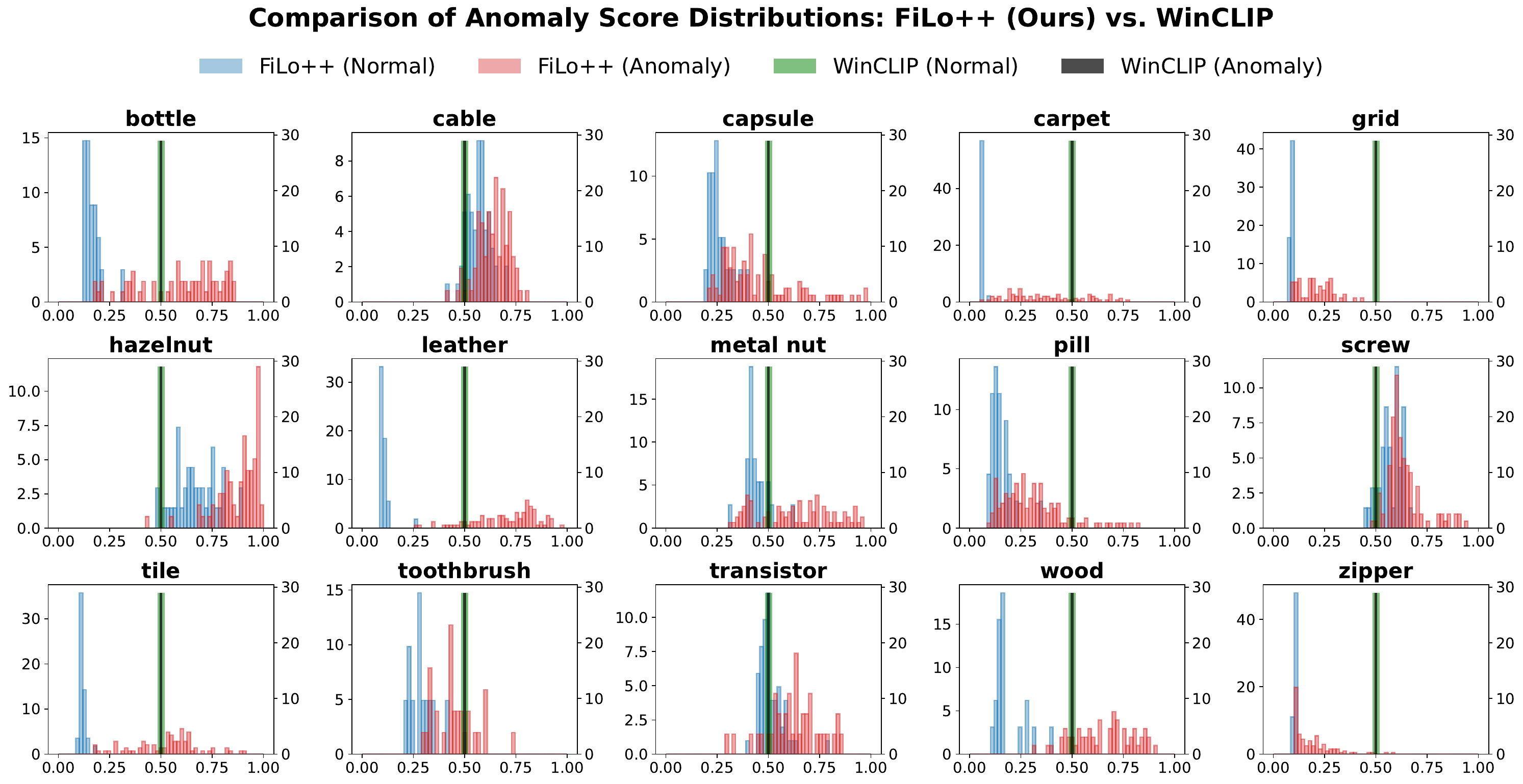} 
  \caption{Comparison of anomaly score distributions between FiLo++ and WinCLIP on the MVTec-AD dataset. The histograms illustrate the separation between normal (blue/green) and anomalous (red/black) samples for each object category. FiLo++ demonstrates significantly better separability compared to WinCLIP.}
  \label{fig:score_map}
\end{figure*}

\begin{figure}[]
  \centering
  \includegraphics[width=0.95\linewidth]{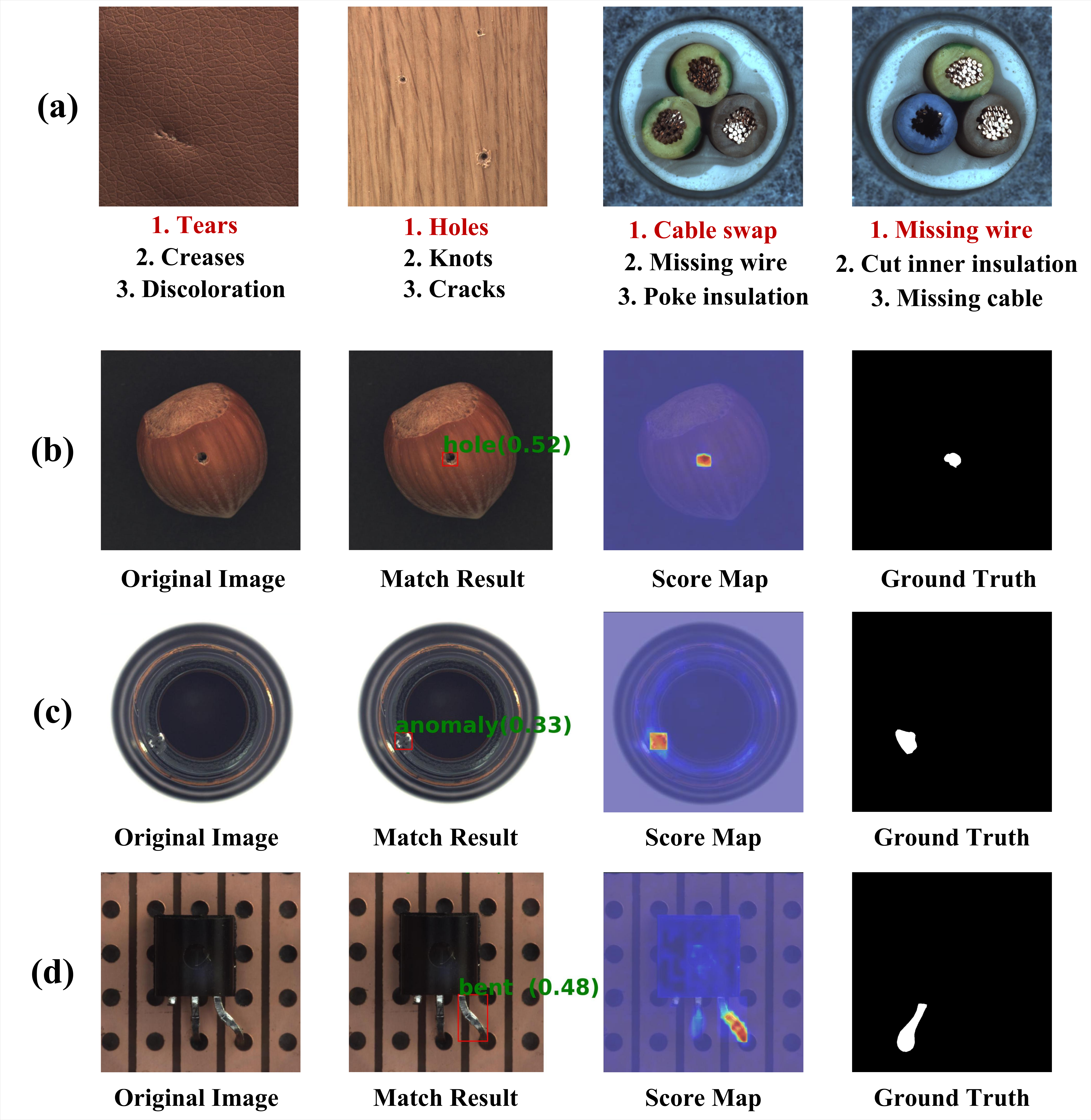} 
  \caption{Qualitative analysis of detection details and failure cases. (a) shows the top-3 most similar text descriptions for the input images. (b) demonstrates the precise alignment between the detected region and the fine-grained text description ("hole"). (c) illustrates a case where the LLM failed to enumerate the specific defect, yet FiLo++ still localized it using the generic "anomaly" prompt. (d) presents a failure case where a missing pin was misclassified as background.}
  \label{fig:detail_match}
\end{figure}

\subsubsection{Adaptively Learned Text Templates}
In Table~\ref{tab:coop} and Fig.~\ref{fig:num_vec}, we analyze the parameters of learnable text vectors. Comparing CoOp~\cite{zhou2022learning} and CoCoOp~\cite{zhou2022conditional}, Table~\ref{tab:coop} shows that CoOp performs better without class names, consistent with AnomalyCLIP~\cite{zhou2023anomalyclip}. Conversely, CoCoOp benefits from including class names. This aligns with CoCoOp's instance-specific nature, which parallels FiLo++'s goal of deriving precise, image-aligned descriptions. Additionally, Fig.~\ref{fig:num_vec} demonstrates that setting the number of learnable vectors to 12 yields optimal performance for both detection and localization.

\subsubsection{Utilization of V-V Attention}
Standard CLIP's QKV attention tends to aggregate features from semantically unrelated regions, frequently causing noisy activations outside object boundaries~\cite{li2023clip}. While recent works like AnoVL~\cite{deng2023anovl} and AnomalyCLIP~\cite{zhou2023anomalyclip} adopt V-V attention to improve feature alignment, we observed that V-V attention can be unstable during training, where minor errors often lead to model collapse. To balance performance and stability, FiLo++ concurrently utilizes both mechanisms. We systematically investigated distinct post-processing strategies and found that applying a simple linear transformation to QKV outputs while feeding V-V outputs into the MDCI module yields the optimal performance for both detection and localization, as evidenced in Table~\ref{tab:qkvvv}.

\subsubsection{Utilization of Different LLMs}
In Table~\ref{tab:llm}, we present a comparative analysis of the model's performance in generating detailed anomaly descriptions utilizing different large language models. The experimental results indicate that the choice of LLM exerts only a marginal influence on the overall performance, suggesting that employing different LLMs to generate comprehensive anomaly descriptions yields largely consistent outcomes.

\subsubsection{Convolution Kernel's Shape of MDCI}
We conduct extensive experiments to investigate the impact of various kernel shapes utilized in the MMCI module. Beginning with the exclusive use of 1×1 convolutional kernels, we incrementally incorporate additional kernel shapes, including 3×3, 5×5, 7×7, 1×5, 5×1, and 9×9, and systematically evaluate the resulting performance, as illustrated in Figure~\ref{fig:mdci_kernel}. Based on our experimental findings, we ultimately select a combination of kernel shapes comprising 1×1, 3×3, 5×5, 7×7, 1×5, and 5×1. This selection effectively leverages the benefits of multi-scale and multi-shape kernels, thereby facilitating precise localization of anomalous regions with diverse sizes and morphologies.

\subsubsection{Impact of Hyperparameter $\lambda$}

In the DefLoc module, we utilize a hyperparameter $\lambda$ to suppress the anomaly scores of regions outside the bounding boxes detected by Grounding DINO. To analyze the sensitivity of $\lambda$, we evaluate the performance on MVTec-AD and VisA datasets with $\lambda$ varying from 0 to 1, as shown in Fig.~\ref{fig:lambda}. From the results, we draw three key observations. First, the value of $\lambda$ has a negligible impact on the Image-level AUC, indicating that global anomaly detection is robust to this suppression. Second, for Pixel-level AUC, settings with $\lambda < 1$ consistently outperform the setting with $\lambda = 1$ (which is equivalent to not using Grounding DINO). This demonstrates that suppressing background noise outside the detected objects effectively reduces false positives and enhances localization accuracy. Third, setting $\lambda = 0$ results in a performance drop compared to intermediate values. This suggests that complete suppression is too aggressive, as it relies entirely on Grounding DINO's localization, which may occasionally be imperfect. Based on these findings, we set $\lambda = 0.4$ in our final implementation.

\subsubsection{Efficiency Analysis}

We further evaluate the computational overhead introduced by the deformable convolution in the DefLoc module. Table~\ref{tab:efficiency} compares the computational overhead of Deformable versus Standard Convolution on an NVIDIA RTX A6000 with $518 \times 518$ resolution. Deformable Convolution incurs only a marginal increase in inference time (+32ms) and memory (+257MB). Given the significant localization improvements demonstrated in Table~\ref{tab:defloc}, this slight additional cost is well-justified for practical applications.

\subsubsection{Impact of the Number of Normal Samples}

To investigate how the number of normal reference samples ($K$) affects the few-shot anomaly detection performance, and to determine whether performance continues to improve or plateaus, we extended our experiments to include 8-shot and 16-shot settings. The results on MVTec-AD and VisA datasets are illustrated in Fig.~\ref{fig:num_k}.
We observe a significant performance boost in both Image-level and Pixel-level AUC as $K$ increases from 1 to 4, demonstrating that a small increase in reference data greatly helps the model estimate the normal distribution. However, the performance gain begins to diminish as $K$ further increases to 8 and 16. Specifically, on the VisA dataset, the Pixel-level AUC shows a clear plateau after 8 shots. This trend suggests that while providing more normal samples is beneficial, our method is highly data-efficient and can achieve near-optimal performance with a relatively small number of samples, after which the marginal returns decrease.

\subsection{Visualization Results}

Fig.~\ref{fig:visual} presents a visual comparison between FiLo++ and leading few-shot anomaly detection methods, specifically PatchCore~\cite{roth2022towards} and AnomalyGPT~\cite{gu2024anomalygpt}. As observed in the second and third rows, PatchCore and AnomalyGPT tend to generate anomaly maps with noticeable noise in normal background regions or fail to precisely delineate the anomaly boundaries. In contrast, FiLo++ demonstrates superior localization capabilities. Even in the zero-shot setting, FiLo++ effectively suppresses background noise and accurately highlights the defects. The 1-shot results further enhance this precision, yielding anomaly maps that are highly consistent with the ground truth.

In Fig.~\ref{fig:score_map}, we present a comparative analysis of anomaly scores generated by WinCLIP~\cite{jeong2023winclip} and FiLo++ for each class object in the MVTec-AD dataset. This combined visualization allows for a direct comparison of the discriminative capability of both methods. As observed in the figure, WinCLIP's score distributions for normal and anomalous samples exhibit severe overlap and are often concentrated within a narrow range, indicating a struggle to effectively distinguish between normal and anomalous samples. In sharp contrast, FiLo++ shows a clear separation: the scores for normal samples are significantly suppressed towards lower values, while those for anomalous samples are pushed towards higher values. This distinct separation, achieved with consistent axis ranges across comparisons, strongly validates the effectiveness of FiLo++'s fine-grained descriptions and adaptively learned text templates in reducing cross-semantic ambiguity.

In Fig.~\ref{fig:detail_match}, we provide a comprehensive qualitative analysis to demonstrate the interpretability, generalization capability, and limitations of FiLo++. 
First, to validate the alignment between detected regions and text descriptions, Fig.~\ref{fig:detail_match}(a) lists the top-3 fine-grained descriptions with the highest similarity scores for the given images. Fig.~\ref{fig:detail_match}(b) further visualizes this correspondence: the model not only detects the anomaly but also accurately grounds it to the specific description "hole" with a high confidence score, verifying the semantic consistency of our method.
Second, we address the concern that LLM-generated descriptions might miss novel anomaly types. As shown in Fig.~\ref{fig:detail_match}(c), although the specific defect was not included in the LLM's enumeration list, our model successfully localized the defect by matching it with the learnable generic "anomaly" template. This indicates that FiLo++ retains strong open-set generalization capabilities and is not strictly bound by the completeness of the LLM's knowledge.
Finally, we analyze a failure case to understand the model's limitations. In Fig.~\ref{fig:detail_match}(d), the model failed to detect a "bent/missing pin" on a component. The false negative occurred because the region where the pin was missing visually resembled the texture of the background board. In a zero-shot setting without a reference image for comparison, the model struggled to distinguish the "absence of an object" from the background. This suggests that incorporating reference-based comparison is crucial for detecting logical anomalies like missing components.

\section{Conclusion}
In this paper, we propose FiLo++, a zero-shot and few-shot anomaly detection framework that addresses the dual challenges of precise detection and accurate localization. By leveraging the FusDes module, which combines the knowledge of large language models with both fixed and learnable text prompts, FiLo++ effectively adapts to diverse anomaly types. The DefLoc module further refines localization through Grounding DINO, position-enhanced text descriptions, and a multi-scale deformable cross-modal interaction module to better handle anomalies of varying shapes and sizes. Additionally, a position-enhanced patch matching strategy boosts few-shot performance by focusing on suspicious regions during inference. Extensive experimental results on MVTec-AD and VisA datasets demonstrate that FiLo++ outperforms existing approaches in both zero-shot and few-shot scenarios, highlighting the potential of integrating powerful language models with advanced vision techniques for anomaly detection.

In future work, we plan to extend the FiLo++ framework to broader applications beyond static anomaly detection. The core components of FiLo++ demonstrate strong potential for transferability to other domains. Specifically, the FusDes module, which leverages compositional prompts to describe complex object states, shares a similar philosophy with compositional action recognition~\cite{yan2023progressive, qu2025learning} and group activity analysis~\cite{yan2020higcin}, where complex activities are understood through their fine-grained atomic components. Furthermore, our proposed position-enhanced patch matching strategy aligns with recent advances in few-shot action recognition~\cite{qu2025mvp}, suggesting its applicability to temporal domains for video anomaly detection. Finally, the cross-modal interaction mechanism in DefLoc can be generalized to other multimodal localization tasks, such as dense audio-visual event localization~\cite{xing2024locality}, by adapting the deformable attention to handle audio-visual correspondences.

{
    \bibliographystyle{IEEEtran}
    \bibliography{main}
}


\vfill

\end{document}